\documentclass[letterpaper,10 pt,conference]{ieeeconf}
\IEEEoverridecommandlockouts                              

\usepackage{amsmath}

\usepackage{array}

\usepackage{pdfpages}

\usepackage{cite}

\usepackage{graphicx}

\usepackage{url}

\begin{document}


\title{\LARGE \bf LightEMMA: A Longitudinal Evaluation of Vision-Language \\ Models for Autonomous Driving}

% \title{\LARGE \bf LightEMMA: A Longitudinal Study of Scaling Effects in Vision-Language Models for Autonomous Driving}


\author{Zhijie Qiao\textsuperscript{1,†}, Haowei Li\textsuperscript{1,†}, Zhong Cao\textsuperscript{1}, Henry X. Liu\textsuperscript{1,2,*}
\thanks{This research was partially funded by the DARPA TIAMAT Challenge (HR0011-24-9-0429).}%
\thanks{$^{1}$Z.~Qiao, H.~Li, Z.~Cao, and H.~X.~Liu are with the Department of Civil and Environmental Engineering, University of Michigan, Ann Arbor, MI 48109, USA.} 
\thanks{$^{2}$H.~X.~Liu is also with the University of Michigan Transportation Research Institute, Ann Arbor, MI 48109, USA. }
\thanks{†These authors contributed equally to this work.}%
\thanks{*Corresponding author: Henry X. Liu (henryliu@umich.edu).}%
}

% \author{Anonymous Authors
% \thanks{Funding information omitted for double-blind review.}
% \thanks{$^{1}$Affiliation omitted for double-blind review.} 
% \thanks{$^{2}$Affiliation omitted for double-blind review.}
% \thanks{$^{3}$Affiliation omitted for double-blind review.}
% \thanks{$^{4}$Affiliation omitted for double-blind review.}
% \thanks{$^{5}$Affiliation omitted for double-blind review.}
% \thanks{†Equal contribution.}%
% \thanks{*Corresponding author information omitted.}%
% }


\maketitle

\begin{abstract}

Rapid advances in vision-language models (VLMs) have generated growing interest in their application to autonomous driving. A prevailing assumption is that successive VLM generations will continually improve driving performance and eventually outperform state-of-the-art methods. To systematically examine this assumption, we introduce LightEMMA, a longitudinal framework for evaluating the autonomous driving performance of VLMs. LightEMMA uses a lightweight, unified evaluation protocol that assesses each model's intrinsic driving capability without model-specific fine-tuning, architectural changes, or prompt engineering. Using this protocol, we evaluate 15 models from five major families on the challenging nuScenes prediction benchmark. Empirical findings show that, despite increased model scale and enhanced general reasoning capabilities, successive VLM generations do not consistently achieve better driving performance. Further analysis of driving scenarios reveals recurring failure modes, including overreliance on historical actions and difficulty reconciling conflicting visual cues. These findings highlight the need for domain-specific adaptation to improve the safety of VLM-based autonomous driving systems. The source code is available at https://github.com/michigan-traffic-lab/LightEMMA.

\end{abstract}


\section{Introduction}

Autonomous vehicles (AVs) have evolved from modular, rule-based architectures \cite{buehler2009darpa, Yang2018} to learning-based frameworks that map raw sensor data to planned trajectories \cite{Hu_2023_CVPR, 9561904, Zeng_2019_CVPR, 9157239, hu2022stp3}. While end-to-end approaches provide a unified pathway from perception to driving decisions, they remain constrained by limited interpretability \cite{10614862} and poor generalization to rare, safety-critical long-tail scenarios \cite{liu2024curse}. Vision-language models (VLMs), equipped with open-world reasoning and contextual scene understanding, have emerged as a promising solution to bridge these gaps. Despite rapid advances in frontier VLM reasoning, their efficacy in autonomous driving remains an open question.

EMMA \cite{hwang2025emma} demonstrated the viability of a unified vision-language approach to autonomous navigation by fine-tuning Gemini \cite{comanici2025gemini} on multimodal sensor and trajectory logs. However, its proprietary implementation limits broader research use by the community. OpenEMMA \cite{Xing_2025_WACV} explored an open-source alternative using a prompt-based framework; yet its evaluation was restricted to a small set of models, offering limited insight into autonomous driving performance across different VLM families and generations.

\begin{figure}[t!]
    \centering
    \vspace{0.2cm}
    \includegraphics[width=1.0\columnwidth]{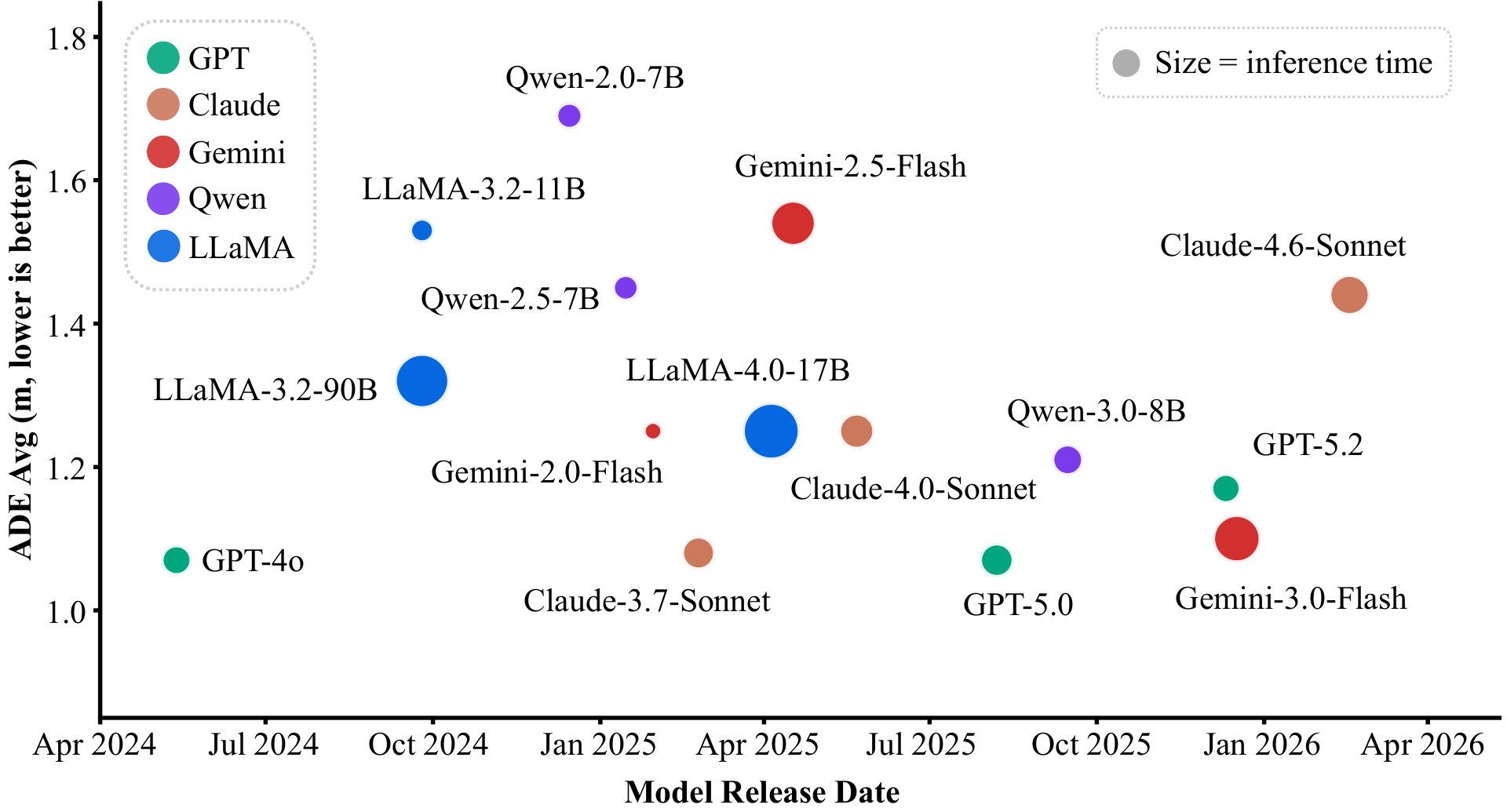}
    \caption{Longitudinal evaluation of VLMs for autonomous driving. Each bubble represents a model evaluated on the nuScenes prediction benchmark and is colored by model family. Bubble size is proportional to average inference time per frame, with larger bubbles corresponding to slower models. Lower ADE reflects better trajectory prediction accuracy.}
    \label{fig:longitudinal}
    \vspace{-0.2cm}
\end{figure}

To address these evaluation gaps, we present LightEMMA, a lightweight framework that supports state-of-the-art and emerging VLMs. Using LightEMMA, we conduct the first longitudinal evaluation of VLMs on the nuScenes prediction benchmark~\cite{9156412}, focusing on inference time and trajectory prediction accuracy (Fig.~\ref{fig:longitudinal}). Under a unified setting without model-specific fine-tuning, architectural changes, or prompt engineering, we assess three years of model development across five major families, enabling direct comparison of their intrinsic driving capabilities. Our results show that newer models with stronger general reasoning do not necessarily achieve better driving performance, underscoring the need for domain-specific adaptation. We hope LightEMMA will serve as a practical evaluation benchmark to support and facilitate future research in VLM-based autonomous driving.

\begin{figure*}[ht!]
    \centering
    \vspace{0.2cm}
    \includegraphics[width=1.0\textwidth]{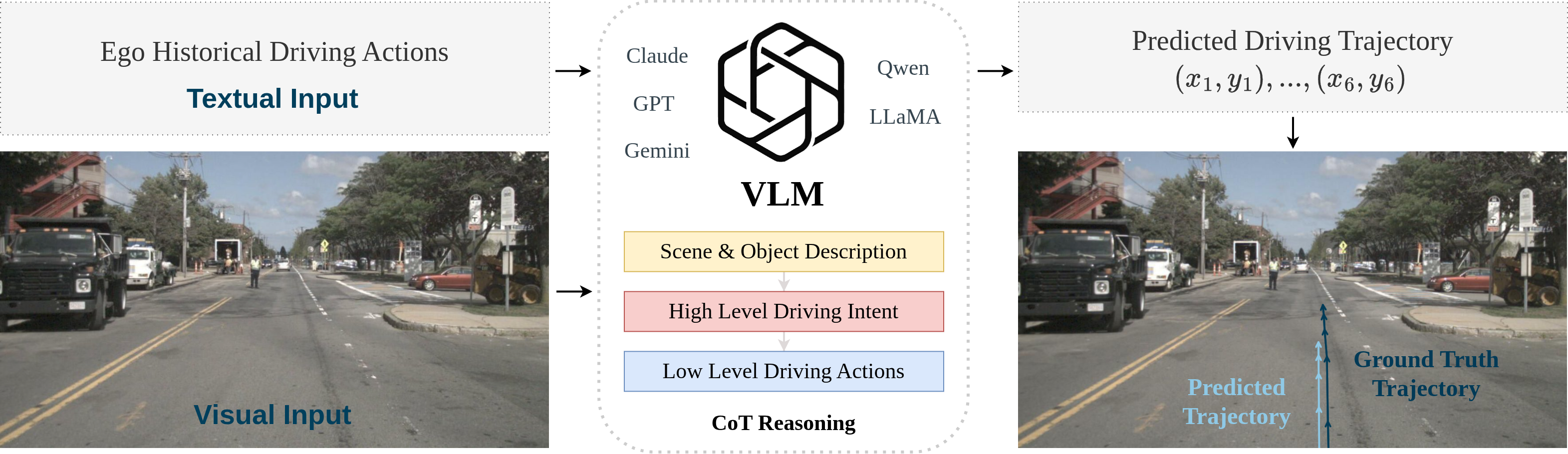}
    \caption{LightEMMA architecture.}
    \label{fig:architecture}
\end{figure*}

The main contributions of this work are as follows:

\begin{enumerate}
\item We present LightEMMA, a lightweight, unified framework for evaluating the autonomous driving capabilities of evolving VLMs.

\item We conduct a longitudinal evaluation spanning five major model families over three years of development. The results show that successive VLM generations do not consistently improve driving performance.

\item Our scenario-based analysis identifies recurring failure modes in common driving situations, highlighting the need for domain-specific adaptation.
\end{enumerate}

\section{Related Work}

\subsection{VLM-Based Driving Systems}

VLM-based driving systems employ natural-language representations to interpret driving scenes, reason about them, and generate vehicle control commands. DriveGPT4 \cite{10629039} adapts LLaMA-2 \cite{touvron2023llama2} to predict next-step speed and turning angle from front-view monocular video and ego speed. The model is fine-tuned and evaluated on BDD-X \cite{kim2018textual}, a dataset pairing driving videos with vehicle controls and textual explanations. DriveGPT4 further augments BDD-X with ChatGPT-generated driving-specific question-answer pairs for visual instruction tuning and flexible human--vehicle interaction. DOLPHINS \cite{10.1007/978-3-031-72995-9_23} extends OpenFlamingo \cite{awadalla2023openflamingo} into a conversational driving assistant using visual inputs, text instructions, and driving history. The model adapts to new driving tasks using prompt examples and corrects errors based on textual feedback without updating parameter weights. DriveMLM \cite{cui2025drivemlm} fine-tunes the Husky-7B VLM \cite{liu2023interngpt} as the behavior-planning module in a closed-loop driving system. The model takes encoded multi-view camera and LiDAR observations, traffic rules, and user commands as input and predicts discrete speed and path decisions accompanied by textual explanations. DriveLM \cite{sima2024drivelm} introduces Graph VQA, which connects perception, prediction, and planning through dependent question-answer pairs. Its DriveLM-Agent jointly performs graph-based reasoning and end-to-end driving on benchmarks derived from nuScenes and CARLA \cite{dosovitskiy2017carla}. LMDrive \cite{shao2024lmdrive} follows natural-language navigation commands and responds to passenger warnings in closed-loop end-to-end driving. Its LLaMA-based model \cite{llama} predicts future waypoints and instruction completion from camera and LiDAR features, while the accompanying LangAuto benchmark tests multi-step and misleading commands in safety-critical scenarios.

\subsection{Efficient and Hybrid VLM Systems}

The computational demands of VLM inference remain a key challenge for real-time deployment. To address this, Gopalkrishnan et al. \cite{gopalkrishnan2024multiframe} introduce EM-VLM4AD, which uses gated attention to combine encoded features from six camera views before feeding them to a compact T5 model \cite{raffel2020exploring} for driving-scene question answering. The model requires less than one-tenth of the memory and computation of other driving VLMs evaluated in the study. DriveVLM \cite{tian2024drivevlm} proposes DriveVLM-Dual, a slow-fast architecture coupling a VLM-based pipeline with a conventional driving pipeline. The VLM branch analyzes complex scenes and generates trajectories at low frequency, while the conventional branch provides 3D spatial grounding and refines these trajectories at high frequency.

\subsection{Driving and Reasoning Benchmarks}

nuScenes~\cite{9156412} contains 1,000 annotated urban scenes with surround-view cameras, radar, and LiDAR for 3D detection, tracking, and motion prediction. NuPrompt~\cite{wu2025language} extends nuScenes with object-centric language descriptions for tracking referenced 3D objects across views and frames. WOD-E2E~\cite{xu2025wod} targets long-tail end-to-end driving using surround-view images, routing information, and ego states; its Rater Feedback Score compares predicted trajectories with human preferences. Reason2Drive~\cite{reason2drive} provides over 600,000 video-text pairs organized into perception, prediction, and reasoning chains, together with a chain-level evaluation metric.

\section{Methodology}

Fig.~\ref{fig:architecture} presents an overview of the LightEMMA architecture. For each inference, the VLM receives the front-view image and the vehicle's driving history. A chain-of-thought (CoT) prompt \cite{wei2022chain} guides the model to produce a sequence of predicted control actions. These actions are numerically integrated to generate the predicted trajectory, which is then compared with the ground-truth trajectory to compute prediction errors. All VLMs are assessed using a uniform prompting and evaluation protocol without model-specific adaptations.

\subsection{VLM Selection}

We evaluate state-of-the-art VLMs from both open-source and commercial offerings across five model families: GPT~\cite{singh2025openai}, Claude~\cite{anthropic2026claude46sonnet}, Gemini~\cite{comanici2025gemini}, Qwen~\cite{bai2025qwen3}, and LLaMA~\cite{llama}. For each family, we select three generational variants to balance model capability, inference time, and cost. This setup enables both cross-family and cross-generational comparisons under identical conditions (Tables~\ref{tab:efficiency} and~\ref{tab:performance}). LLaMA did not support vision-language capabilities prior to version 3.2, and subsequent releases still offer limited multimodal options, constraining the available generational variants. Commercial models are accessed via paid APIs, while open-source models are downloaded from Hugging Face and deployed locally with H100 GPUs. The minimum number of GPUs required for each model is reported in Table~\ref{tab:efficiency}.

\subsection{Camera Input}

In our framework, the VLM receives raw front-view camera images directly, without preprocessing or auxiliary perception overlays \cite{yolov8, vidal2026hybrid}. Our experiments show that overlaying external perception cues (e.g., bounding boxes) yields no performance gain because VLMs can already resolve fine-grained scene elements directly from raw images. Furthermore, we provide only the current driving frame rather than concatenating multiple past frames \cite{Xing_2025_WACV, 10629039}, as additional frames offer negligible performance gains while substantially increasing inference time and cost. Video-based architectures with specialized temporal encodings \cite{sun2019videobert, NEURIPS2022_416f9cb3} may better capture motion dynamics but constitute a separate research direction.

\subsection{Driving History Input}

Our framework encodes the ego vehicle's driving history as six actions, each represented by a speed-curvature pair \((v, \kappa)\), where speed describes longitudinal motion and curvature describes lateral motion. The actions are sampled at 0.5-second intervals to match the temporal resolution of the nuScenes dataset and cover a total duration of 3 seconds. Compared with Cartesian coordinates \((x, y)\), this representation provides a more intuitive description of vehicle dynamics. The historical sequence, together with the current front-view camera image, is used as input to the CoT prompting procedure detailed in the next subsection.

\subsection{VLM Prompting}

Following established prompting strategies in prior driving studies \cite{hwang2025emma, tian2024drivevlm}, we adopt a structured Chain-of-Thought (CoT) pipeline to guide the VLM through hierarchical scene reasoning and action generation. The intermediate output of each stage is sequentially fed into subsequent prompts, providing cumulative context across three stages:

\begin{enumerate}
    \item \textbf{Scenario Description:} The VLM receives the front-view camera image and describes the overall scene, including lane markings, traffic lights, vehicles, pedestrians, and other relevant elements.
    
    \item \textbf{High-Level Driving Intent:} The generated scene description is combined with the ego vehicle's historical driving actions, enabling the VLM to interpret its past behavior in the current scene and predict the next high-level driving action.
    
    \item \textbf{Low-Level Driving Commands:} The scene description and generated high-level driving action are provided to the VLM to produce a structured sequence of low-level driving commands formatted as [\((v_1, \kappa_1)\), \((v_2, \kappa_2)\), \((v_3, \kappa_3)\), \((v_4, \kappa_4)\), \((v_5, \kappa_5)\), \((v_6, \kappa_6)\)], where each pair specifies the speed and curvature for one timestep, without supplementary text or explanations.
\end{enumerate}

\subsection{Trajectory Generation}

We generate the predicted trajectory by propagating future vehicle states through kinematic integration in an egocentric reference frame, starting from the initial position \((x_0, y_0)=(0,0)\) and orientation \(\theta_0=0\). Given a sequence of driving commands \(\{(v_t, \kappa_t)\}\) produced by the VLM, where \(v_t\) denotes vehicle speed and \(\kappa_t\) denotes curvature at time step \(t\), we iteratively update the vehicle's position and orientation.

Using a fixed time step of \(\Delta t=0.5\,\text{s}\), consistent with the sampling interval of the nuScenes dataset, the vehicle state is updated as follows:
\begin{align}
x_{t+1} &= x_t + v_t \cos(\theta_t)\,\Delta t, \\[6pt]
y_{t+1} &= y_t + v_t \sin(\theta_t)\,\Delta t, \\[6pt]
\theta_{t+1} &= \theta_t + \kappa_t v_t\,\Delta t.
\end{align}

The resulting trajectory consists of six predicted positions sampled at \(0.5\,\text{s}\) intervals over a \(3\,\text{s}\) prediction horizon.

\subsection{Example Workflow}

To illustrate the prompting workflow, we present the model outputs for the scenario shown in Fig.~\ref{fig:architecture} as follows:

\textbf{Scenario Description}: The image shows an urban street scene with active road construction or maintenance work\dots\ A police officer is standing in the middle of the road directing traffic\dots\ There are several construction vehicles visible including a black dump truck parked on the left side of the road along with other work vehicles\dots\ Traffic appears to be controlled through this construction zone with the presence of the officer and possible lane restrictions.

\textbf{High-Level Driving Intent}: Given the current scene with active construction and a police officer directing traffic in the middle of the road, the ego vehicle should significantly decelerate by approximately 2--3 m/s\(^{2}\) to prepare for possible stopping\dots\ The vehicle should continue following the lane while being prepared to come to a complete stop if directed by the officer.

\textbf{Low-Level Driving Commands}: [(6.0, -0.001), (5.0, -0.001), (4.0, 0.0), (3.5, 0.0), (3.0, 0.0), (3.0, 0.0)].

\begin{table*}[t]
\vspace{0.2cm}
\caption{Comparison of Model Efficiency and Computational Cost. $\dagger$ denotes anomalous values discussed in Section~\ref{sec:token_cost}.}
\centering
\begin{tabular}{m{3.0cm}>{\centering\arraybackslash}m{2.15cm}>{\centering\arraybackslash}m{2.15cm}>{\centering\arraybackslash}m{2.15cm}>{\centering\arraybackslash}m{2.15cm}>{\centering\arraybackslash}m{2.15cm}}
\hline
Model & Infer Time (s) & Infer Cost (\textcent) & Input Tokens & Output Tokens & H100 GPUs \\
\hline
GPT-4o & 12.1 & 1.40 & 4402 & 341 & - \\
GPT-5.0 & 15.3 & 0.93 & 3922 & 440 & - \\
GPT-5.2 & 11.6 & 1.68 & 6375 & 400 & - \\
Claude-3.7-Sonnet & 14.8 & 2.47 & 5948 & 461 & - \\
Claude-4.0-Sonnet & 17.0 & 2.63 & 6074 & 540 & - \\
Claude-4.6-Sonnet & 22.2 & 2.76 & 6238 & 660 & - \\
Gemini-2.0-Flash & \textbf{4.5} & \textbf{0.07} & 6398 & 272 & - \\
Gemini-2.5-Flash & 28.5 & 1.35\textsuperscript{$\dagger$} & 1968\textsuperscript{$\dagger$} & 402 & - \\
Gemini-3.0-Flash & 31.0 & 0.35 & 4484 & 404 & - \\
Qwen-2.0-VL-7B & 9.1 & - & 6593 & 341 & 1 \\
Qwen-2.5-VL-7B & 8.8 & - & 6554 & 318 & 1 \\
Qwen-3.0-VL-8B & 12.8 & - & 5359 & 375 & 1 \\
LLaMA-3.2-11B & 7.5 & - & 1039\textsuperscript{$\dagger$} & 313 & 1 \\
LLaMA-3.2-90B & 40.8 & - & 1078\textsuperscript{$\dagger$} & 357 & 3 \\
LLaMA-4.0-17B-Scout & 44.6 & - & 1311\textsuperscript{$\dagger$} & 546 & 3 \\
\hline
\end{tabular}
\label{tab:efficiency}
\end{table*}

\section{Experiments}

Using the proposed framework, we evaluate the selected VLMs on the nuScenes trajectory prediction task across 150 test scenarios comprising 3,908 frames. Our evaluation covers inference time, token usage and cost, model response reliability, and trajectory prediction accuracy.

\subsection{Inference Time}

Table~\ref{tab:efficiency} summarizes inference times across all evaluated models, reporting the average processing time per frame for the complete CoT procedure. Gemini-2.0-Flash achieves the fastest inference at 4.5 seconds per frame. Its successors, despite being lightweight variants, are substantially slower. The Qwen family demonstrates consistently low latency across all generations, owing to its smaller model scale. In contrast, the LLaMA family exhibits the highest latency overall, with both the 90B and 17B-Scout variants requiring over 40 seconds per frame. GPT models maintain relatively stable inference times across generations, while Claude models show steady increases across successive releases (rising from 14.8 to 22.2 seconds). In general, newer generations are slower than their predecessors, indicating that expanded capabilities come at the expense of speed. Even the fastest model falls far short of real-time requirements, highlighting the need for strategies such as distillation \cite{tang2019distillingtaskspecificknowledgebert} or dual-system designs \cite{tian2024drivevlm}.

\subsection{Token Usage and Cost}
\label{sec:token_cost}

We report the average input and output token counts per frame using each model's official token-counting procedure. Input token counts substantially exceed output tokens due to image encoding. For commercial APIs, costs are calculated by cross-referencing billing records with token usage and official pricing schedules. All costs in Table~\ref{tab:efficiency} are reported in cents per frame.

Across most models, input token counts range from 4,000 to 6,000 per frame, while output token counts remain in the low hundreds. Among commercial offerings, Claude is the most expensive, though its billing remains consistent and predictable across successive releases. Gemini is the least expensive family overall. However, Gemini-2.5-Flash exhibits anomalously low input token counts and billing inconsistencies, which are subsequently resolved in Gemini-3.0-Flash. Similarly, the LLaMA family reports unexpectedly low input token counts across all releases, likely due to the omission of image tokens from its official tokenizer.

\begin{table*}[t]
\vspace{0.2cm}
\caption{Comparison of Model Performance on nuScenes Prediction Task.}
\centering
\begin{tabular}{m{3.0cm}>{\centering\arraybackslash}m{1.1cm}>{\centering\arraybackslash}m{1.5cm}>{\centering\arraybackslash}m{1.5cm}>{\centering\arraybackslash}m{1.5cm}>{\centering\arraybackslash}m{1.5cm}>{\centering\arraybackslash}m{1.6cm}>{\centering\arraybackslash}m{1.2cm}}

\hline
Model & FE (\%) & FE Corr (\%) & ADE 1s (m) & ADE 2s (m) & ADE 3s (m) & ADE avg (m) & FDE (m) \\
\hline
GPT-4o & 7.78 & 0.08 & \textbf{0.28} & \textbf{0.92} & \textbf{2.01} & \textbf{1.07} & \textbf{2.34} \\
GPT-5.0 & 0.0 & 0.0 & 0.28 & 0.92 & 2.03 & 1.07 & 2.36 \\
GPT-5.2 & 0.0 & 0.0 & 0.30 & 1.02 & 2.19 & 1.17 & 2.54 \\
Claude-3.7-Sonnet & 0.0 & 0.0 & 0.28 & 0.94 & 2.04 & 1.08 & 2.36 \\
Claude-4.0-Sonnet & 0.03 & 0.0 & 0.30 & 1.07 & 2.39 & 1.25 & 2.78 \\
Claude-4.6-Sonnet & 0.56 & 0.0 & 0.35 & 1.25 & 2.73 & 1.44 & 3.17 \\
Gemini-2.0-Flash & 0.7 & 0.0 & 0.31 & 1.08 & 2.36 & 1.25 & 2.79 \\
Gemini-2.5-Flash & 0.03 & 0.0 & 0.37 & 1.32 & 2.93 & 1.54 & 3.42 \\
Gemini-3.0-Flash & 0.0 & 0.0 & 0.28 & 0.93 & 2.08 & 1.10 & 2.44 \\
Qwen-2.0-VL-7B & 0.03 & 0.03 & 0.64 & 1.61 & 2.83 & 1.69 & 3.16 \\
Qwen-2.5-VL-7B & 0.0 & 0.0 & 0.47 & 1.33 & 2.55 & 1.45 & 2.90 \\
Qwen-3.0-VL-8B & 1.25 & 0.0 & 0.38 & 1.09 & 2.16 & 1.21 & 2.49 \\
LLaMA-3.2-11B & 0.69 & 0.03 & 0.52 & 1.42 & 2.67 & 1.53 & 3.03 \\
LLaMA-3.2-90B & 0.0 & 0.0 & 0.35 & 1.15 & 2.46 & 1.32 & 2.86 \\
LLaMA-4.0-17B-Scout & 0.05 & 0.0 & 0.29 & 1.06 & 2.40 & 1.25 & 2.81 \\
\hline
\end{tabular}
\label{tab:performance}
\end{table*}

\subsection{Response Reliability}

In the final inference stage, we observed several types of format errors, as summarized by the format error (FE) rate in Table~\ref{tab:performance}. Although each VLM was instructed to return outputs strictly in the format [\((v_1, \kappa_1)\), \((v_2, \kappa_2)\), \((v_3, \kappa_3)\), \((v_4, \kappa_4)\), \((v_5, \kappa_5)\), \((v_6, \kappa_6)\)] without extraneous text, occasional deviations still occurred. These included missing brackets or commas, explanatory text or punctuation, and incorrect output lengths. Representative failure examples and parsing logs are provided in our open-source release.

Since all models were evaluated using identical prompts and workflows, we attribute these failures to inherent model stochasticity rather than a systematic issue in our framework. To address errors that impede downstream analysis, we implemented a simple error-handling technique that attempts to extract twelve values from each output and organize them into six speed-curvature pairs. The corrected results, reported in the FE Corr. column of Table~\ref{tab:performance}, show that all models achieve format error rates below 0.1\%, demonstrating that the vast majority of outputs are valid for analysis. Nevertheless, the occurrence of these errors reflects the models' underlying difficulty in strictly adhering to the prescribed output format.

\subsection{Trajectory Prediction Accuracy}

Prediction accuracy is evaluated using official nuScenes metrics: Average Displacement Error (ADE) at 1, 2, and 3 seconds, and Final Displacement Error (FDE). For frames in which format errors persist after correction, we rerun inference until a valid prediction is obtained, ensuring all 3,908 frames are evaluated for every model.

Table~\ref{tab:performance} presents the evaluation results, with cross-family and cross-generational trends further illustrated in Fig.~\ref{fig:longitudinal}. GPT-4o achieves the lowest ADE and FDE values, with GPT-5.0 and Claude-3.7-Sonnet following closely behind. Gemini-3.0-Flash also performs competitively as the strongest variant in the Gemini family. Qwen-3.0-VL-8B delivers remarkably strong results for a model of its scale, rivaling much larger commercial offerings. The LLaMA family performs the worst overall, exhibiting the highest errors across all horizons.

Beyond individual model comparisons, Fig.~\ref{fig:longitudinal} captures the broader longitudinal trend: successive VLM generations do not consistently improve autonomous driving performance. Several newer models exhibit higher ADE than their predecessors within the same family, while larger model scale does not reliably guarantee better driving performance. These findings underscore a fundamental disconnect between general-purpose VLM progress and driving competence, highlighting the critical need for domain-specific adaptation.

\section{Scenario-Based Failure Analysis}

While metrics such as ADE and FDE quantify average displacement error, they do not explain underlying reasoning failures or contextual misinterpretations \cite{zhai2023ADMLP, li2024ego}. To gain deeper insight into model behavior, we examine six representative scenarios depicted in Fig.~\ref{fig:examples}. Each subfigure provides an illustrative comparison between VLM-predicted and ground-truth trajectories.

\subsection*{Case 1: Trajectory Bias from Historical Actions}

Fig.~\ref{fig:examples}.1 illustrates a scenario in which the ground-truth trajectory continues straight, while the predicted trajectory incorrectly indicates a strong right turn and fails to account for an obstacle on the right. Although initially counterintuitive, this behavior is frequently observed across models. The error occurs because the AV had completed a right turn at an intersection immediately before this frame, causing the historical actions to retain a pronounced rightward curvature. However, the models struggle to adjust their predictions based on the current front-view image and mistakenly project the preceding turn forward. Shortly after the vehicle resumes a straight path, the models correctly adjust and begin predicting straight trajectories again. Similar errors also occur following left turns.

\subsection*{Case 2: Insufficient Context from Visual Cues}

\begin{figure*}[ht]
    \centering
    \vspace{0.2cm}
    \includegraphics[width=\textwidth]{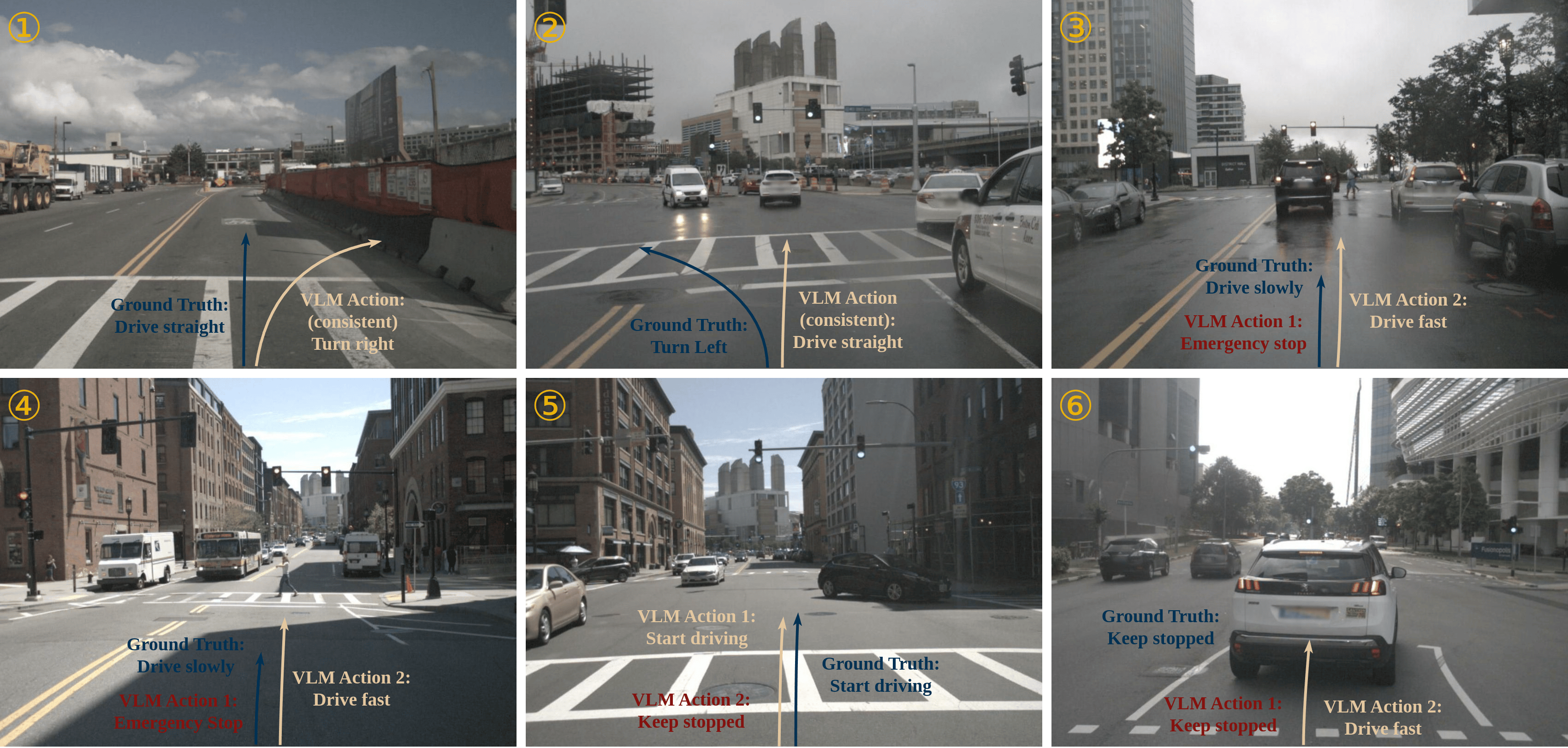}
    \caption{Representative driving scenarios illustrating common failure modes across six cases (1--6). Each subfigure displays the front-view camera image alongside the corresponding predicted and ground-truth trajectories over a 3-second horizon.}
    \label{fig:examples}
\end{figure*}

Fig.~\ref{fig:examples}.2 presents a scenario in which the ground-truth trajectory turns left, while the models incorrectly predict continuing straight. Although the scenario is challenging because it lacks explicit left-turn markings or dedicated traffic signals, implicit cues remain. For instance, the AV occupies the leftmost lane, while vehicles in the adjacent lane to the right are positioned to continue straight. Handling such cases reliably may require additional contextual information, such as explicit navigation instructions.

\subsection*{Cases 3 \& 4: Divergent Responses to Stop Signals}

Fig.~\ref{fig:examples}.3 illustrates a scenario in which the AV gradually approaches a stopped vehicle at an intersection controlled by a red traffic signal. The ground-truth trajectory shows the AV smoothly decelerating until it comes to a complete stop behind the leading vehicle. The model predictions fall into two distinct categories, neither of which accurately matches the ground-truth behavior. The first category, generally associated with lower ADE, correctly identifies the red traffic signal and the stopped vehicle ahead, yet predicts excessively abrupt braking rather than controlled, gradual deceleration. Although these models recognize critical visual cues, they lack the spatial reasoning needed to estimate distance and modulate their response accordingly. Their behavior appears event-triggered, responding immediately to the red light without accounting for the developing scenario. The second category, generally associated with higher ADE, predicts that the AV will continue through the intersection without slowing or stopping, effectively ignoring both the stationary vehicle and the red traffic signal. A similar pattern appears in Fig.~\ref{fig:examples}.4, where models again either predict an abrupt stop or overlook the pedestrian and red signal entirely.

\subsection*{Case 5: Divergent Responses to Green Signals}

Fig.~\ref{fig:examples}.5 depicts a scenario in which the AV is initially stationary at a signal-controlled intersection. When the signal changes from red to green, the ground-truth trajectory shows the AV promptly accelerating and smoothly traversing the intersection. Models with lower ADE closely replicate this behavior, recognizing the green signal as an indication to proceed and predicting appropriate acceleration trajectories. Conversely, models with higher ADE remain stationary and fail to associate the green signal with acceleration, highlighting their difficulty in translating dynamic visual cues into timely control actions. In practice, such delayed responses could obstruct traffic flow and increase the risk of rear-end collisions.

\subsection*{Case 6: Model Responses to Conflicting Visual Cues}

The final example, shown in Fig.~\ref{fig:examples}.6, presents a scenario in which even models with low ADE exhibit different behaviors. As in Fig.~\ref{fig:examples}.5, the traffic signal has just changed from red to green. One group of models recognizes the green light and predicts immediate acceleration, overlooking the vehicle directly ahead. Another group correctly reconciles the conflicting cues and predicts that the AV must remain stationary because the vehicle ahead blocks its path despite the green signal. This scenario requires resolving competing scene elements, which remains challenging even for stronger models. These divergent responses highlight the variability of VLM decision-making in driving tasks and the need for additional safety mechanisms, such as redundant perception checks or rule-based fallbacks.

\section{Conclusion}

In this work, we presented LightEMMA, a lightweight, unified framework for evaluating VLMs in autonomous driving. Our longitudinal evaluation across five model families shows that newer generations and stronger general reasoning do not necessarily improve driving performance, highlighting the need for domain-specific adaptation. Our scenario-based failure analysis further reveals recurring limitations in spatial reasoning and dynamic scene understanding. We hope that LightEMMA will support continued evaluation of VLMs and advance their application to autonomous driving.

\bibliographystyle{ieeetr}
\bibliography{LightEMMA}

\end{document}